\documentclass[runningheads]{llncs}
\usepackage{subcaption}
\usepackage{graphicx}

\usepackage{multirow}
\usepackage{amsmath,amssymb,amsfonts}

\usepackage{bbding}
\usepackage{hyperref}
\hypersetup{hypertex=true,
colorlinks=true,
linkcolor=blue,
anchorcolor=blue,
citecolor=blue}

\usepackage{caption}
\begin{document}
\title{Exploring Unsupervised Cell Recognition with Prior Self-activation Maps}
%
%

\author{Pingyi Chen\inst{1,2,3}
\and
Chenglu Zhu \inst{2,3}\and
Zhongyi Shui\inst{1,2,3}\and
Jiatong Cai \inst{2,3}\and
Sunyi Zheng\inst{2,3}\and
Shichuan Zhang\inst{1,2,3}\and
Lin Yang\inst{2,3}}

\authorrunning{P. Chen et al.}

\titlerunning{Exploring Unsupervised Cell Recognition with Prior Self-activation Map}
%

\institute{College of Computer Science and Technology, Zhejiang University \and
Research Center for Industries of the Future, Westlake University\and School of Engineering, Westlake University.
\\
\email{\{chenpingyi,yanglin\}@westlake.edu.cn}}

\maketitle              
\begin{abstract}
The success of supervised deep learning models on cell recognition tasks relies on detailed annotations.
Many previous works have managed to reduce the dependency on labels. However, considering the large number of cells contained in a patch, costly and inefficient labeling is still inevitable. 
To this end, we explored label-free methods for cell recognition. Prior self-activation maps (PSM) are proposed to generate pseudo masks as training targets.  
To be specific, an activation network is trained with self-supervised learning. The gradient information in the shallow layers of the network is aggregated to generate prior self-activation maps. Afterward, a semantic clustering module is then introduced as a pipeline to transform PSMs to pixel-level semantic pseudo masks for downstream tasks. 
We evaluated our method on two histological  datasets: MoNuSeg (cell segmentation) and BCData (multi-class cell detection). Compared with other fully-supervised and weakly-supervised methods, our method can achieve competitive performance without any manual annotations. Our simple but effective framework can also achieve multi-class cell detection which can not be done by existing unsupervised methods. The results show the potential of PSMs that might inspire other research to deal with the hunger for labels in medical area. 

\end{abstract}
\section{Introduction}
\label{sec:introduction}
Cell recognition serves a key role in  exploiting pathological images for disease diagnosis. Clear and accurate cell shapes provide rich details: nucleus structure, cell counts, and cell density of distribution. 
Hence, pathologists are able to conduct a reliable diagnosis according to the information from the segmented cell, which also improves their experience of routine pathology workflow
\cite{elston1991pathological,le1989prognostic}.

In recent years, the advancement of deep learning has facilitated significant success in medical images 
\cite{naylor2018segmentation,liu2019nuclei,2015unet}. 
However, the supervised training requires massive manual labels,
especially when labeling cells in histopathology images.  A large number of cells are required to be labeled, which results in inefficient and expensive annotating processes. It is also difficult to achieve accurate labeling because of the large variations among different cells and the variability of reading experiences among pathologists.

Work has been devoted to reducing dependency on manual annotations recently. Qu et al. use  points  as supervision \cite{qu2019weakly}. It is still a labor-intensive task due to the large number of objects contained in a pathological image. With regard to unsupervised cell recognition, traditional methods can segment the nuclei by clustering or morphological processing. But these methods suffer from worse performance than deep learning methods. Among AI-based methods, some works use domain adaptation to realize unsupervised instance segmentation  \cite{2019ddmrl,sifa,cycada}, which transfers the source domain containing annotations to the unlabeled target. However, their satisfactory performance depends on the appropriate annotated source dataset.  Hou et al. \cite{hou2019robust} proposed to synthesize training samples with GAN. It relies on predefined nuclei texture and color. Feng et al. \cite{feng2021mutual} achieved unsupervised detection and segmentation by a mutual-complementing network. It combines the advantage of correlation filters and deep learning but needs iterative training and finetuning. 

CNNs with inductive
biases have priority over local features  of the nuclei with
dense distribution and semi-regular shape. In this paper, we proposed a simple but effective framework for unsupervised cell recognition. Inspired by the strong representation capability of self-supervised learning, we devised the prior self-activation maps (PSM) as the supervision for downstream cell recognition tasks. Firstly, the activation network is initially trained with self-supervised learning like predicting  instance-level contrastiveness. Gradient information accumulated in the shallow layers of the activation network is then calculated and aggregated with the raw input information.  These features extracted from the activation network are then clustered to generate pseudo masks which are used for downstream cell recognition tasks. In the inferring stage, the networks which are supervised by pseudo masks are directly applied for cell detection or segmentation. To evaluate the effectiveness of PSM, we evaluated our method on two datasets. Our framework achieved comparable performance on cell detection and segmentation on par with supervised methods. Code is available at \href{https://github.com/cpystan/PSM}{https://github.com/cpystan/PSM}.

\section{Method}
\begin{figure*}

\includegraphics[width=\textwidth]{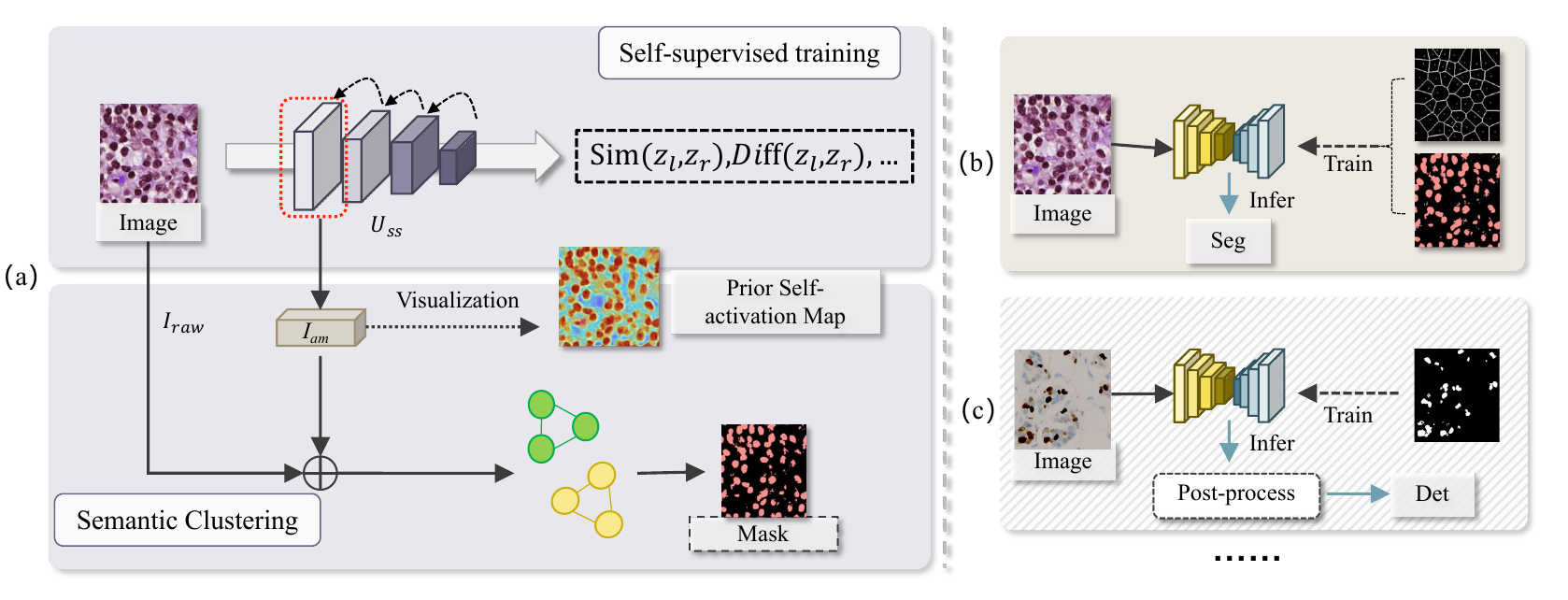}

\caption{The framework of our proposed model. (a) The top block shows the activation network which aggregates gradient information to generate self-activation maps. The bottom block presents the work process of semantic clustering. (b-c) The right part is the cell detection and the cell segmentation network which are supervised by the pseudo masks.  }  
\label{frame}
\end{figure*}

The structure of our proposed method is demonstrated in Fig. \ref{frame}. 
Firstly, an activation network $U_{ss}$ is trained with self-supervised learning. After the backpropagation of gradients,  gradient-weighted features are exploited to generate the self-activation maps (PSM). 
Next is semantic clustering where the PSM is combined with the raw input to generate pseudo masks.
These pseudo masks can be used as supervision for downstream tasks. Related details are discussed in the following.

\subsection{Proxy Task}
We introduce self-supervised learning to encourage the network to focus on the  local features in the image.  And our experiments show that neural networks are capable of adaptively recognizing nuclei with dense distribution and semi-regular shape.  Here, we have experimented with several basic proxy tasks below.

\textbf{ImageNet pre-training:}  It is  straightforward to exploit the models pre-trained on natural images. In this strategy, we directly extract the gradient-weighted feature map in the ImageNet pre-trained network and generate prior self-activation maps. 

\textbf{Contrastiveness:} Following the contrastive learning \cite{chen2020simple} methods, the network is encouraged to distinguish between different patches. For each image, its augmented view will be regarded as the positive sample, and the other image sampled in the training set is defined as the negative sample. The network is trained to minimize the distance between positive samples. It also maximizes the distance between the negative sample and the input image. The optimization goal can be denoted as:
\begin{equation}
    L_{dis}(Z_l,Z_r,Z_n)=diff(Z_l,Z_r)-diff(Z_l,Z_n),
\end{equation}

where $L_{dis}$ is the loss function. $Z_l$, $Z_r$, and $Z_n$ are representations of the input sample, the positive sample, and the negative sample, respectively. In addition, $diff(\cdot)$ is  a function that measures the difference of embeddings.

\textbf{Similarity:} LeCun et al. \cite{siamese} proposed a Siamese network to train the model with a similarity metric. We also adopted a weight-shared network to learn the similarity discrimination task. In specific, the pair of samples (each input and its augmented view) will be fed to the network, and then embedded as high-dimensional vectors $Z_l$ and $Z_r$ in the high-dimensional space, respectively. 
Based on the similarity measure $sim(\cdot)$, $L_{dis}$ is introduced to reduce the distance, which is denoted as,
\begin{equation}
    L_{dis}(Z_l,Z_r)=-sim(Z_l,Z_r) = diff(Z_l,Z_r). 
\end{equation}
Here, maximizing the similarity of two embeddings is equal to minimizing their difference. 

\subsection{Prior Self-activation Map}
The self-supervised model $U_{ss}$ is constructed by sequential blocks which contain several convolutional layers, batch normalization layers, and activation layers. 
The self-activation map of a certain block can be obtained by nonlinearly mapping the weighted feature maps $A^k$ :
\begin{equation}
I_{am}= ReLU(\sum_k {\alpha}_k A^k),
\end{equation}
where $I_{am}$ is the prior self-activation map. $A^k$ indicates the $k$-th feature map in the selected layer. ${\alpha}_k$ is the weight of each feature map, 
which is defined by global-average-pooling the gradients of output $z$ with regard to $A^k$:
\begin{equation}
    {\alpha}_k = \frac{1}{N} \sum_i \sum_j \frac{\partial {z}}{\partial {A_{ij}^k}},
\end{equation}
where $i$,$j$ denote the height and width of output, respectively, and $N$ indicates the input size. The obtained features are visualized in the format of the heat map which is later transformed to pseudo masks by clustering.

\subsubsection{Semantic Clustering.}
 We construct a semantic clustering module (SCM) which converts prior self-activation maps to pseudo masks. 
In SCM, the original information is included to strengthen the detailed features.
It is defined as:
\begin{equation}
    {I_{f}}= I_{am} + \beta \cdot I_{raw},
\label{detail}
\end{equation}
where ${I_{f}}$ denotes the fused semantic map, $I_{raw}$ is the raw input,  $\beta$ is the weight of $I_{raw}$. 

To generate semantic labels, an unsupervised clustering method K-Means is selected to directly split all pixels into several clusters and obtain foreground and background pixels. 
Given the semantic map $I_f$ and its $N$ pixel features $F=\{f_i | i\in \{1,2,...N\}\}$, $N$ features are partitioned into $K$ clusters $S = \{S_i | i \in \{1,2,...K\}  \}$, $K<N$. The goal is to find $S $ to reach the minimization of within-class variances as follows:
\begin{equation}
    min \sum_{i=1}^K \sum_{f_j\in S_i} ||f_j - c_i||^2,
\end{equation}
where $c_i$ denotes the centroid of each cluster $S_i$. After clustering, the pseudo mask $I_{sg}$ can be obtained. 

\subsection{Downstream Tasks}
In this section, we introduce the training and inferring of cell recognition models. 

\subsubsection{Cell Detection.}
For the task of  cell detection, a detection network  is trained under the supervision of pseudo mask $I_{sg}$. 
In the inferring stage, the output of the detection network is a score map. Then, it is post-processed to obtain the detection result. 

The coordinates of cells can be got by searching local extremums in the score map, which is described below:
\begin{equation}
T_{(m,n)}=\left\{
\begin{array}{l}
1, \quad p_{(m,n)}>p_{(i,j)},\quad {\forall}(i,j)\in D_{(m , n)},\\
0, \quad otherwise,
\end{array}
\right.
\end{equation}
where $T_{(m,n)}$ denotes the predicted  label at location of $(m,n)$, $p$ is the value of the score map and $D_{(m , n)}$ indicates the neighborhood of point $(m,n)$.  $T_{(m,n)}$ is exactly the detection result.

\subsubsection{Cell Segmentation.}

Due to the lack of instance-level supervision, the model does not perform well in distinguishing adjacent objects in the segmentation.
To further reduce errors and uncertainties, the Voronoi map $I_{vor}$ which can be transformed from $I_{sg}$ is utilized to encourage the model to focus on instance-wise features. In the Voronoi map, the edges are labeled as background and the seed points are denoted as foreground. Other pixels are ignored.

We train the segmentation model with these two types of labels. 
The training loss function can be formulated as below,
\begin{equation}
\begin{split}
L=  \lambda [ y \log I_{vor} &+ (1-y) \log (1-I_{vor})] \\
                                  &+ (1 - y)\log (1-I_{sg}),
                                  \end{split}
\label{vor}
\end{equation}
where $\lambda$ is the partition enhancement coefficient. In our experiment, we discovered that false positives  hamper the effectiveness of  segmentation due to the ambiguity of cell boundaries. Since that, only the background of $I_{sg}$ will be concerned to eliminate the influence of false positives in instance identification.

\section{Experiments}
\subsection{Implementation Details}
\subsubsection{Dataset.} We validated the proposed method on the public dataset of Multi-Organ Nuclei Segmentation (MoNuSeg) \cite{2017monu} and Breast tumor Cell Dataset (BCData) \cite{2020BCData}.
MoNuSeg consists of 44 images of size 1000 $\times$ 1000 with around 29,000 nuclei boundary annotations. 
BCData is a public large-scale breast tumor dataset containing 1338 immunohistochemically Ki-67 stained images of size 640 $\times$ 640.

\subsubsection{Evaluation Metrics.}
In our experiments on MoNuSeg, F1-score and IOU are employed to evaluate the segmentation performance. Denote $TP$, $FP$, and $FN$ as the number of true positives, false positives, and false negatives. Then F1-score and IOU can be defined as: $F1=2TP/(2TP+FP+FN)$, $IOU=TP/(TP+FP+FN)$. In addition, common object-level indicators such as Dice coefficient and Aggregated Jaccard Index (AJI) \cite{2017monu} are also considered to assess the segmentation performance.

In the experiment on BCData, precision (P), recall (R), and F1-score are used to evaluate the detection performance. Predicted points will be matched to  ground-truth points one by one. And those unmatched points are regarded as false positives. Precision and recall are: $P=TP/(TP+FP)$, and $R=TP/(TP+FN)$. In addition, we introduce MP and MN to evaluate the cell counting results. 'MP' and 'MN' denote the mean average error of positive and negative cell numbers.

\subsubsection{Hyperparameters.}
Res2Net101 \cite{res2net} is adopted as the activation network $U_{ss}$ with random initialization of parameters. The positive sample is augmented by rotation. The weights $\beta$ are set to  $2.5$ and $4$ for training in MoNuSeg and BCData, respectively.  The weight $\lambda$ is 0.5. The analysis for $\beta$ and  $\lambda$ is included in the supplementary. Pixels
of the fused semantic map will be decoupled into three piles
by K-Means.
. The following segmentation and detection are constructed with  ResNet-34. They are optimized using CrossEntropy loss by the Adam optimizer for $100$ epochs with the initial learning rate of $1e^{-4}$. The function $diff(\cdot)$
is instantiated as the measurement of Manhattan distance.

\begin{table}[t] 
  \centering
  \caption{Results on MoNuSeg. According to the requirements of each method, various labels are used: localization (Loc) and contour (Cnt). * indicates the model is trained from scratch with the same hyperparameter as ours.}
  \resizebox{0.85\linewidth}{!}
{
  \begin{tabular}{l|cc|cc|cc}
   \hline
  \multirow{2}{*}{Methods} & \multirow{2}{*}{Loc} & \multirow{2}{*}{Cnt} & \multicolumn{2}{c|}{Pixel-level} & \multicolumn{2}{c}{Object-level} \\
  \cline{4-5} \cline{6-7}
  ~ & ~ & ~ & IoU & F1 score & Dice & AJI \\
  \hline
    Unet*\cite{2015unet} & \CheckmarkBold & \CheckmarkBold & 0.606 & 0.745 & {0.715} & 0.511 \\
    MedT\cite{2021medt} & \CheckmarkBold & \CheckmarkBold & {0.662} & {0.795} & - & - \\
    CDNet \cite{he2021cdnet} & \CheckmarkBold & \CheckmarkBold & - & - & 0.832 & 0.633 \\
    Competition Winner \cite{2017monu} & \CheckmarkBold & \CheckmarkBold & - & - & - & 0.691 \\
    \hline

    Qu et al.\cite{qu2019weakly}& \CheckmarkBold & \XSolidBrush & 0.579 & 0.732 & 0.702 & 0.496 \\
    Tian et al.\cite{tian2020weakly}& \CheckmarkBold & \XSolidBrush & 0.624 & 0.764 & 0.713 & 0.493 \\
    \hline
        CellProfiler \cite{carpenter2006cellprofiler} & \XSolidBrush & \XSolidBrush & - & 0.404 & 0.597 & 0.123 \\
            Fiji\cite{fiji} & \XSolidBrush & \XSolidBrush & - & 0.665 & 0.649 & 0.273 \\
    CyCADA\cite{cycada} & \XSolidBrush & \XSolidBrush & - & 0.705 & - & 0.472 \\
    Hou et al. \cite{hou2019robust} & \XSolidBrush & \XSolidBrush & - & 0.750 & - & 0.498 \\
    \hline
    \textbf{Ours} & \XSolidBrush & \XSolidBrush & 0.610 & 0.762 & 0.724 & {0.542} \\
  \hline
  \end{tabular}}

\label{tab:comp}
\end{table}

\begin{table}[th]
  \centering
  \caption{Results on BCData. According to the requirements of each method, various labels are used: localization (Loc) and the number (Num) of cells.}
\resizebox{0.9\linewidth}{!}
{
 \begin{tabular}{l | cc | c | ccc|c|c}
  \hline
  Methods & Loc & Num & Backbone & P & R & F1 score&MP$\downarrow$& MN$\downarrow$\\
  \hline
    CSRNet\cite{csrnet} & \CheckmarkBold & \CheckmarkBold & ResNet50 &0.824 & 0.834& 0.829  &9.24 &24.90\\
    SC-CNN\cite{snpcnn} & \CheckmarkBold & \CheckmarkBold & ResNet50 &0.770 & 0.828& 0.798   & 9.18 &20.60\\
    U-CSRNet\cite{2020BCData} & \CheckmarkBold & \CheckmarkBold & ResNet50 & 0.869  & 0.857    &   0.863     & 10.04 & 18.09 \\
    TransCrowd\cite{transcrowd}& \XSolidBrush & \CheckmarkBold & Swin-Transformer&-&-&- & 13.08 &33.10\\
    \hline
    \textbf{Ours} & \XSolidBrush & \XSolidBrush & ResNet34& 0.855& 0.771 & 0.811& 8.89 &28.02\\
  \hline
  \end{tabular}}

\label{tab:comp1}
\end{table}

\subsection{Result}
\begin{figure*}[t]
\includegraphics[width=\textwidth]{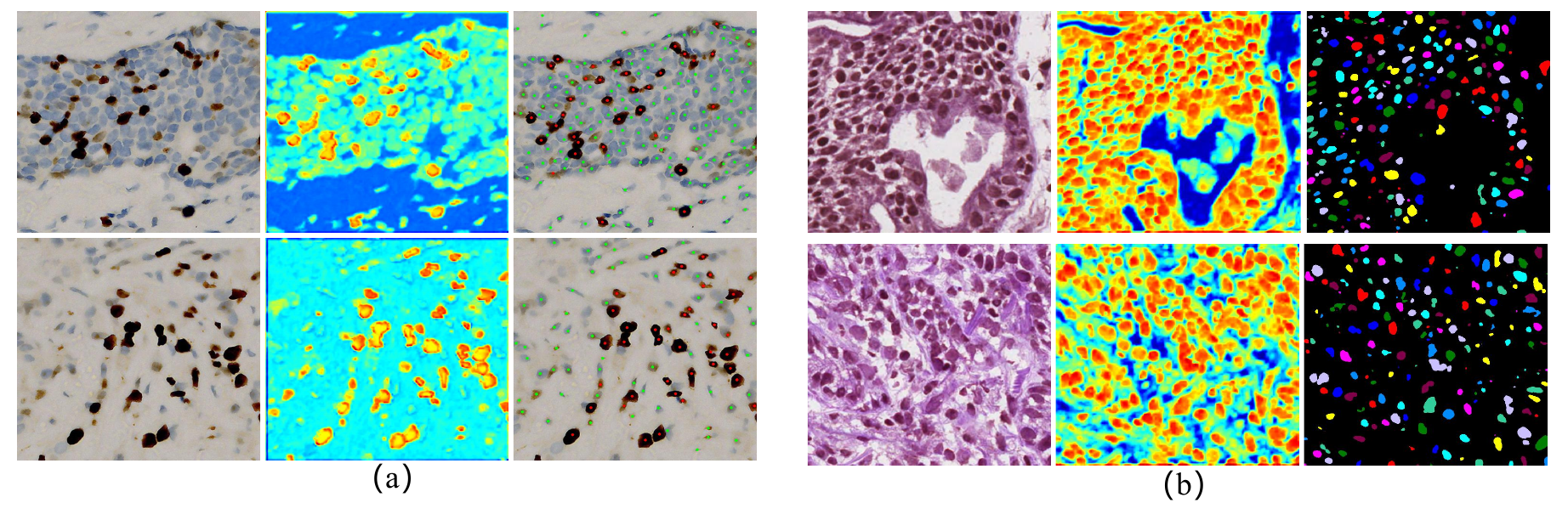}

\caption{Visualization. (a) Typical results of multi-class detection, the red dot and green dot represent positive and negative cells respectively. (b) Typical results of cell segmentation.  } \label{result} 
\end{figure*}

This section includes the discussion of results which are visualized in Fig. \ref{result}

\subsubsection{Segmentation.}
In MoNuSeg Dataset, four fully-supervised methods Unet \cite{2015unet}, MedT \cite{2021medt}, CDNet \cite{he2021cdnet}, and the competition winner \cite{2017monu}  are adopted to estimate the upper limit as shown in the first four rows of Table \ref{tab:comp}.
Two weakly-supervised models trained with only point annotations are also adopted as the comparison. Compared with the method \cite{tian2020weakly} fully exploiting localization information, ours can achieve better performance without any annotations in object-level metrics (AJI).
In addition, two unsupervised methods using traditional image processing tools \cite{fiji,carpenter2006cellprofiler} and two unsupervised methods \cite{cycada,hou2019robust} with deep learning are compared.
Our framework has achieved promising performance because robust low-level features are exploited to generate high-quality pseudo masks.

\subsubsection{Detection.}
Following the benchmark of BCData, metrics of detection and counting are adopted to evaluate the performance as shown in Table \ref{tab:comp1}.
The first three methods are fully supervised methods which predict probability maps to achieve detection.

Furthermore, TransCrowd \cite{transcrowd} with the backbone of Swin-Transformer is employed as the weaker supervision trained by cell counts regression. 
By contrast, even without any annotation supervision, compared to CSRNet \cite{csrnet}, NP-CNN \cite{snpcnn} and U-CSRNet \cite{2020BCData}, our proposed method still achieved comparable performance. Especially in terms of MP, our model surpasses all the baselines. It is challenging to realize multi-class recognition in an unsupervised framework. Our method still achieves not bad counting results on negative cells. 

\subsubsection{Ablation Study.}
\begin{figure}[t]
    \centering
	\begin{minipage}{0.45\linewidth}
		\centering
		
		\setlength{\abovecaptionskip}{0.28cm}
		\includegraphics[width=\linewidth]{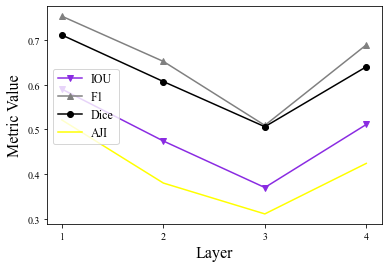}
		\caption{Analysis of the depth of the extracted layer.}
  \label{layers}
	\end{minipage}
	\hfill
	\begin{minipage}{0.45\linewidth}
         \captionof{table}{The training strategy analysis}
		\centering
        \begin{tabular}{c|c|c}
        \hline
        Training Strategy & F1 & AJI \\ \hline
        ImageNet Pretrained & 0.658 & 0.443 \\ \hline
        Similarity & \textbf{0.750} & \textbf{0.542} \\ \hline
        Contrastiveness & 0.741 & 0.498 \\ \hline
        \end{tabular}

        \label{proxy}
	    \end{minipage}
\end{figure}

Ablation experiments are built on MoNuSeg.
In our pipeline, the activation network can be divided into four layers which consists of multiple basic units including ReLU, BacthNorm, and Convolution.  We exploit the prior self-activation maps generated from different depths in the model after training with the same proxy tasks. As shown in Fig. \ref{layers}, the performance goes down and up with we extracting features from deeper layers. Due to the relatively small receptive field, the shallowest layer in the activation network is the most capable to translate local descriptions

We have also experimented with different types of proxy tasks in a self-supervised manner, as shown in Table \ref{proxy}. We can see that relying on the pre-trained models with external data can not improve the results of subsequent segmentation. The model achieves similar pixel-level performance (F1) when learning similarity or contrastiveness. But similarity learning makes the model performs better in object-level metrics (AJI) than contrastive learning. The high intra-domain similarity hinders the
comparison between constructed image pairs. Unlike natural
image datasets containing diverse samples, the minor inter
class differences in biomedical images may not fully exploit 
the superiority of contrastive learning.

\section{Conclusion}

In this paper, we proposed the prior self-activation map (PSM) based framework for unsupervised cell segmentation and multi-class detection. The framework is composed of an activation network, a semantic clustering module  
(SCM), and the networks for cell recognition.
The proposed PSM  has a strong capability of learning low-level representations to highlight the area of interest without the need for manual labels. SCM is designed to serve as a pipeline between representations from the activation network and the downstream task. And our segmentation and detection network are supervised by the pseudo masks. In the whole training process, no manual annotation is needed.  Our unsupervised method was evaluated on two publicly available datasets and obtained competitive results compared to the methods with annotations.  In the future, we will apply our PSM to other types of medical images to further release the dependency on annotations.

\textbf{Acknowledgement.} This study was partially supported by the National Natural Science Foundation of China (Grant no. 92270108), Zhejiang Provincial Natural Science Foundation of China (Grant no. XHD23F0201), and the Research Center for Industries of the Future (RCIF) at Westlake University.

\bibliographystyle{splncs04}
\end{document}


\title{Supplementary}
\author{Pingyi Chen\inst{1,2,3}
\and
Chenglu Zhu \inst{2,3}\and
Zhongyi Shui\inst{1,2,3}\and
Jiatong Cai \inst{2,3}\and
Sunyi Zheng\inst{2,3}\and
Shichuan Zhang\inst{1,2,3}\and
Lin Yang\inst{2,3}\textsuperscript{(\Letter)}}

\institute{College of Computer Science and Technology, Zhejiang University \and
Research Center for Industries of the Future, Westlake University\and School of Engineering, Westlake University.
\\
\email{\{chenpingyi,yanglin\}@westlake.edu.cn}}
\maketitle    
%


\titlerunning{Supplementary}
\begin{table}[h]
  \centering
  \caption{The effect of $\lambda$ which indicates the partition enhancement ratio.}
  \scriptsize\begin{tabular}{l | cc | cc}
  \multirow{2}{*}{Method} & \multicolumn{2}{c|}{Pixel-level} & \multicolumn{2}{c}{Object-level} \\
  \cline{2-3} \cline{4-5}
  ~ & IoU & F1 score & Dice & AJI \\
  \hline
    w/o  & 0.501 & 0.722 & 0.623 & 0.390 \\
    \hline
    w/ and ${\lambda}$ = 0.5 &  0.610 & 0.762 & 0.724 & 0.542 \\
    w/ and ${\lambda}$ = 1 & 0.587 & 0.739 & 0.706 & 0.512 \\
    w/ and ${\lambda}$ = 2 & 0.587 & 0.747 & 0.708 & 0.507 \\
    w/ and ${\lambda}$ = 5 & 0.581 & 0.740 & 0.702 & 0.498 \\
    w/ and ${\lambda}$ = 10 & 0.572 & 0.720 & 0.693 & 0.490 \\
  \hline
  \end{tabular}
\label{tab:lamda}
\end{table}

\begin{figure}
\centering
\includegraphics[width=\linewidth]{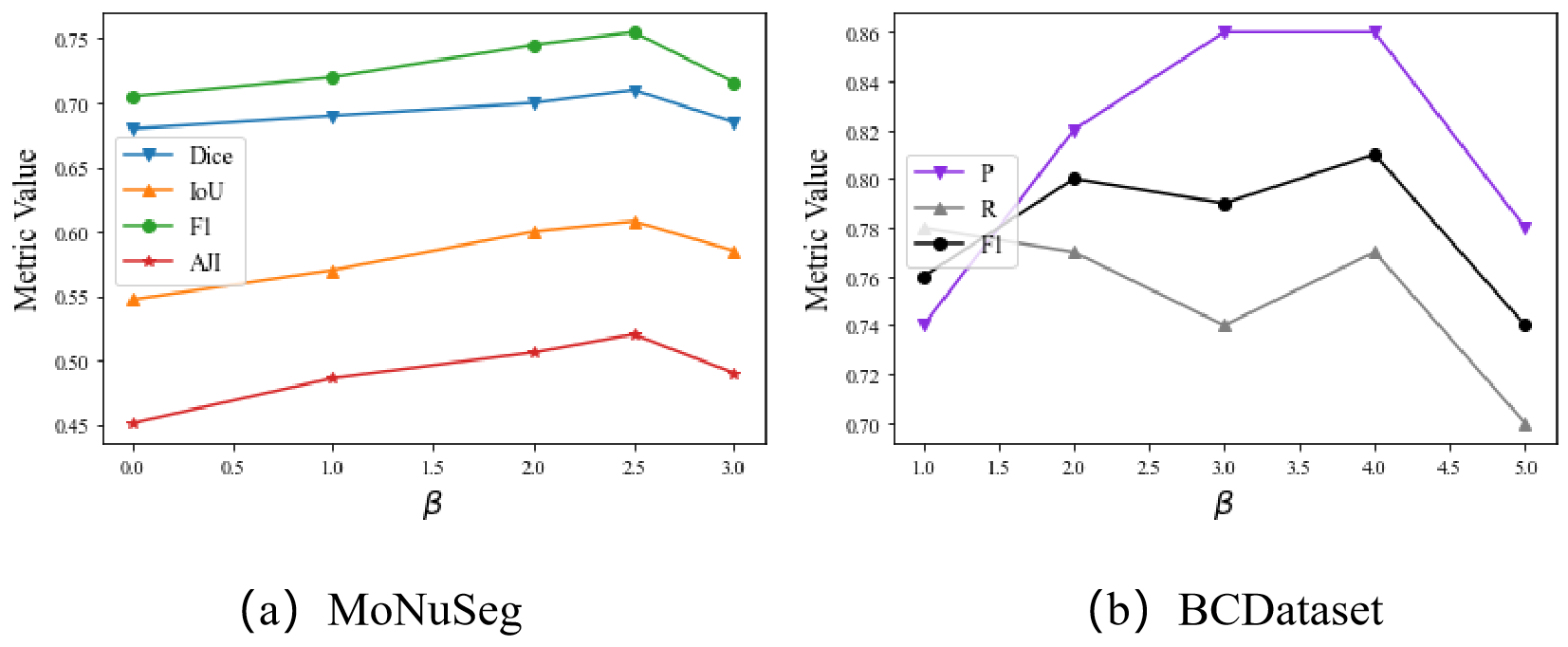}
\caption{The effect of hyperparameter $\beta$ on MoNuSeg and BCData datasets.  }
\label{fig:beta_metric}
\end{figure}